\documentclass[letterpaper, 10 pt, conference]{ieeeconf}  

\IEEEoverridecommandlockouts

\usepackage{cite}
\usepackage{amsmath,amssymb,amsfonts}
\usepackage{algorithmic}
\usepackage{graphicx}
\usepackage{textcomp}
\usepackage{xcolor}

\usepackage{caption}
\usepackage{subcaption}
\usepackage{hyperref}
\usepackage{algorithm}

\usepackage{array}

\usepackage{listings}

\usepackage{booktabs}
\usepackage{multirow}
\usepackage{siunitx}

\def\BibTeX{{\rm B\kern-.05em{\sc i\kern-.025em b}\kern-.08em
    T\kern-.1667em\lower.7ex\hbox{E}\kern-.125emX}}

\title{OceanPlan: Hierarchical Planning and Replanning for Natural Language AUV Piloting in Large-scale Unexplored Ocean Environments}

\author{Ruochu Yang$^{1}$, Fumin Zhang$^{1,2}$, Mengxue Hou$^{3}$
\thanks{This research work is supported by ONR grants N00014-19-1-2556 and N00014-19-1-2266;  AFOSR grant FA9550-19-1-0283; NSF grants GCR-1934836, CNS-2016582 and ITE-2137798; and NOAA grant NA16NOS0120028.}
\thanks{$^{1}$ School of Electrical and Computer Engineering, Georgia Institute of Technology, Atlanta, USA. $^{2}$ Department of Electrical and Computer Engineering, Department of Mechanical and Aerospace Engineering, The Hong Kong University of Science and Technology, Hong Kong, China. $^{3}$ School of Electrical Engineering, University of Notre Dame, Notre Dame, USA.}
}

\begin{document}

\maketitle
\thispagestyle{empty}
\pagestyle{empty}

\begin{abstract}
We develop a hierarchical LLM-task-motion planning and replanning framework to efficiently ground an abstracted human command into tangible Autonomous Underwater Vehicle (AUV) control through enhanced representations of the world. We also incorporate a holistic replanner to provide real-world feedback with all planners for robust AUV operation. While there has been extensive research in bridging the gap between LLMs and robotic missions, they are unable to guarantee success of AUV applications in the vast and unknown ocean environment. To tackle specific challenges in marine robotics, we design a hierarchical planner to compose executable motion plans, which achieves planning efficiency and solution quality by decomposing long-horizon missions into sub-tasks. At the same time, real-time data stream is obtained by a replanner to address environmental uncertainties during plan execution. Experiments validate that our proposed framework delivers successful AUV performance of long-duration missions through natural language piloting.  
\end{abstract}


\section{Introduction}
\label{intro}

\begin{figure*}
        \centerline{\includegraphics[width=\textwidth]{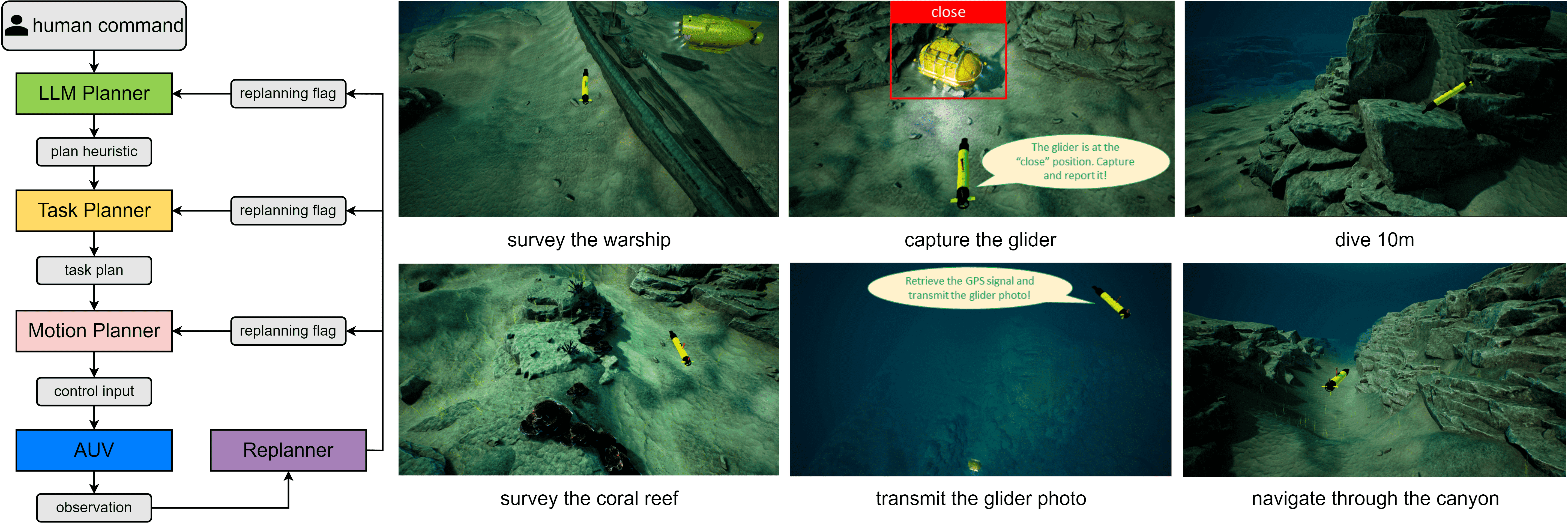}}
        \caption{OceanPlan can accomplish AUV missions through natural language commands in the large-scale unexplored ocean.}
        \label{oceanplan}
\end{figure*}

AUVs have been actively used in a wide range of marine applications like hurricane prediction and ocean observation systems \cite{doi:10.1080, 4476150}. However, it is usually heavy labor to pilot AUVs in real-world missions given complex mechanical manuals and mission files. It would be a relief to simplify this AUV piloting procedure with one single command, preferably in natural language. Recent upsurge of LLMs offers us a promising option to achieve this vision. While LLMs are proved to internalize rich knowledge in text formats, it raises a critical issue of leveraging such knowledge for embodied robot capabilities in the unknown and dynamically changing physical world \cite{tellex2020robots}. We keep reflecting on this question: given an abstracted human command "Search the aborted warship", how can we bridge the gaps between overarching LLMs and physical AUV motions to accomplish the mission? To fulfill our vision, there are inevitable challenges in the realm of marine robotics. Underwater localization is widely recognized as a major challenge, since Global Positioning System (GPS) cannot penetrate seawater. Other terrestrial or aerial localization methods like map-based localization are unavailable in the ocean as well. Another challenge is limited information about the ocean environment. The ocean topography is unknown {\it a priori} and highly unstructured. Geographical terrains and marine objects can pose unexpected collision to AUVs. Moreover, due to the vast spatial scale of the ocean and limited onboard battery capacity of the AUV, it is not enough for the planner to generate only feasible solutions. To ensure mission success, the planner has to offer energy-efficient strategies. 

To address these unique challenges in marine robotics, we design a framework OceanPlan to accomplish efficient and robust AUV missions as shown in Figure~\ref{oceanplan}. In general, our work belongs to the hierarchical planning category where there are many similar works. To the best of our knowledge, this is the first LLM-task-motion planning and replanning framework for natural language piloting in the AUV domain. By labeling a work as missing a (re)planner, we imply that it doesn't explicitly account for that (re)planner. For example, while every robot needs a motion planner to execute skills, some works assume that these skills are ready on hand. \cite{huang2022language} only uses LLMs to directly call pre-defined robotic APIs. \cite{liang2023code} integrates LLM planning with motion planning but no task planning and replanning. \cite{lin2023generalized} implements replanning at both task and motion planning levels without LLMs for human interaction. \cite{singh2023progprompt} incorporates LLM planning to enhance task planning in specific tasks. \cite{huang2022inner} leverages replanning to deal with invalid LLM planning. \cite{huang2023voxposer} develops a framework of LLM-motion planning and replanning for robotic manipulators. \cite{shah2023navigation} encompasses all parts of LLM-task-motion planning, but lack of replanning to handle uncertainty. Other than simply connecting planners, our framework leverages their unique advantages to achieve efficient AUV missions targeted to marine robotics challenges. For robust AUV operation in the uncertain ocean, a holistic replanner coordinates with all planners based on real-time environmental feedback. An underwater simulator is necessary because conducting empirical algorithms or train DRL algorithms on real AUVs is impractical. The main contributions of this work are summarized as follows:
\begin{itemize}
    \item  We present a hierarchical LLM-task-motion planning and replanning framework OceanPlan to pilot AUVs in natural language, specifically for long-horizon missions in large-scale unexplored ocean environments. This hierarchy is key to achieving planning efficiency with pruned search branches and grounding robotic plans with refined representations of the real world. 
   
    \item  We instantiate OceanPlan on a photo-realistic ocean simulator HoloEco featuring various scenes and a polished AUV model EcoMapper. Through comprehensive simulation, we verify efficiency (due to hierarchical architecture) and robustness (due to replanning) of our work targeting inherent AUV challenges.

\end{itemize}

\section{Related Works}
\label{related works}

LLMs have exhibited considerable capabilities of interpreting natural language in the context of real world. Recently, unifying LLMs with robotics has emerged as a rapidly evolving research topic. Many works focus on LLM planning pipelines for robotic execution. One common approach relies on prompt engineering to let LLMs derive a sequential plan towards a user query \cite{huang2022language, singh2023progprompt}. \cite{ahn2022can} harnesses LLM scores and RL-based affordances to select the most provable skill of completing the overall instruction. \cite{liang2023code} utilizes LLMs to create new code polices for unseen scenarios. \cite{dai2023optimal} guarantees efficiency and optimality by LLM semantic guidance and LTL consistent guidance. \cite{yu2023l3mvn} leverages LLMs specifically for navigation tasks in unknown environments. \cite{liu2023llm+} empowers LLMs with classical PDDL planners to achieve optimal planning. \cite{valmeekam2022large} makes a sharp point that LLMs are not really planning because they don't fully understand the physical world and their "plans" depend largely on provided prompts. It is also problematic for LLMs to generate a long-horizon plan at the very beginning without subsequent updates. \cite{rajvanshi2023saynav} proposes dynamical planning to update future actions based on current and visited scenes. Taking these issues into consideration, we resort to the following well-established robotic field.

Task and Motion Planning (TAMP) is a vastly investigated field in the robotics community \cite{garrett2021integrated}. Classical TAMP works are established in deterministic and fully observable space, branching into topics like pick-place planning \cite{gualtieri2021robotic}, manipulation planning \cite{zhang2023multi}, path planning \cite{burke2023learning}, and rearrangement planning \cite{ding2023task}. It is a fundamental extension to consider inevitable uncertainty in the real world \cite{lozano1984automatic}. \cite{curtis2022long} temporally decomposes long-horizon problems into a sequence of short horizons. \cite{hou2023interleaved} develops an interleaved DFS-BnB and MCTS method to achieve low computation cost and plan optimality. In light of potential failures in the real world, closed-loop replanning serves as a suitable solution \cite{garrett2020online}. Works span from re-prompting LLMs with corrective instructions \cite{sharma2022correcting}, receiving real-time environmental feedback \cite{huang2022inner}, leveraging model-based control like MPC \cite{huang2023voxposer}, integrating embodied modalities \cite{driess2023palme}, to hierarchical replanning at both logic and motion levels \cite{lin2023generalized}. Reinforcement Learning (RL) has been closely linked with motion planning, where a control policy is directly trained out of robotic action-reward datasets. \cite{munos2014bandits} illustrates intrinsic connections between RL and planning. \cite{smith2022legged} proposes a real-world RL system for fine-tuning locomotion policies of legged robots. \cite{zhu2017target} guides indoor robot navigation through a DRL policy conditioned on target and current images. \cite{devo2020towards} develops generalized exploration policies over unseen environments by separately training object localization and navigation networks.

\section{Problem Formulation}
\label{problem formulation}

In the 3D ocean world, we denote the AUV physical state as $s_t = [x, y, z, \alpha, \beta, \omega]^T \in S $, where $t$ is the timestep, $x, y, z$ are Cartesian coordinates, $\alpha, \beta, \omega$ are roll-pitch-yaw angles, and $S$ is the state domain. Denote $u_t =  [\phi, \psi]^T \in U$ as the AUV control input at timestep $t$, where $\phi$ is the heading angle in the 2D x-y plane, $\psi$ is the pitch angle to enable vertical movement, and $U$ is the control domain. We denote the real-world observation as $z_t$ obtained from AUV sensors. We denote the human command as $q$, which is an abstracted text specifying AUV missions. Since planning all the way down from the abstracted command $q$ to physical AUV control $u_t$ is extremely long-horizon, we decompose this overall problem into three sub-problems.

\subsection{LLM Planning}
To streamline planning from the robot intelligence level to the human intelligence level, we resort to LLM planning. Since the ocean environment is usually unexplored with open-set objects, we leverage  LLMs' generalized inferring capabilities trained out of vast amounts of information. We denote $h$ as a plan heuristic, which is a textual description interpreted by LLMs about the most probable way of achieving the human command $q$ given semantic representations of the current environment. Therefore, we formulate the following LLM planning sub-problem to command AUVs in natural language and bias the long-horizon search towards this abstracted command in the large-scale unknown ocean.

\textbf{Problem 1 (LLM Planning):} 
Given a human command $q$ and the current observation $z_t$, compute a plan heuristic $h$ which is structured into a symbolic goal state $s_{goal}$ and a symbolic initial state $s_{init}$.

\subsection{Task Planning}
\label{problem task planning}

We consider that the AUV can perform a set of pre-defined actions, each of which is described by a set of preconditions and effects. To describe preconditions and effects using logical predicates, we abstract the AUV physical state $s_t$ into a symbolic state. We also define a set of abstracted actions $\mathcal{A}$. Each action $ \{ a_i \} _{i=1}^N \in \mathcal{A} $ is instantiated with its preconditions and effects and executed by a series of AUV control inputs $u_t$ computed by a control policy $\pi_{a_i}$. Additionally, considering that the LLM planner is essentially a semantics-based planner which struggles to understand the physical world, the plan directly generated by the LLM may not be executable in the real world. Especially when the AUV is navigating in a vast environment, it is impractical to factorize continuous numerical state-action space into textual prompts. Therefore, we formulate the task planning sub-problem as follows. 

\textbf{Problem 2 (Task Planning):}
Given a plan heuristic $h$ and a set of actions $\mathcal{A}$, compute a task plan $ \Pi = \{ a_k \}_{k=1}^T \in \mathcal{A}$, which transitions the initial state $s_{init}$ to the goal state $s_{goal}$.

\subsection{Motion Planning}
\label{problem motion planning}

We rely on motion planning to execute physical control on the AUV in a collision-free manner. Due to the unknown ocean environment, the AUV dynamics will contain uncertainty. Hence we represent the AUV dynamics as an unknown MDP $M = \langle S,U,P \rangle$, where $S$ is a set of states $s_t$, $U$ is a set of control inputs $u_t$, and $P$ is the unknown state transition function. Further because of no localization system under water, the AUV has no access to its true state $s_t$. We assume that the AUV is equipped with an on-board camera providing RGB images as observations $z_t$ of its surrounding environment. Therefore, our goal is to ﬁnd a policy based on the observations, denoted as $\pi(u_t|z_t)$, that maximizes the expected return $ G_t  = \sum_{k=0}^\infty \gamma^k R_{t+k+1}$, where $R_k$ represents the $k$-step reward and $\gamma$ is the discount factor. Since we have no explicit distribution of reward $R$ due to the unknown target position and ocean map, we sample AUV trajectories (action-observation pairs) in the simulator to learn the reward distribution and thus find the optimal policy.

\textbf{Problem 3 (Motion Planning):}
Given an action $a_k$ in the task plan $\Pi$, learn its associated control policy $\pi_{a_k}(\cdot | z_t) $ from the sampled trajectories. Given the current observation $z_t$, compute the control input $u_t$ from the learned control policy $ u_t \sim \pi_{a_k}(\cdot | z_t) $.

\section{Methodology}
\label{methodology}

\begin{figure*}
        \centerline{\includegraphics[width=\textwidth]{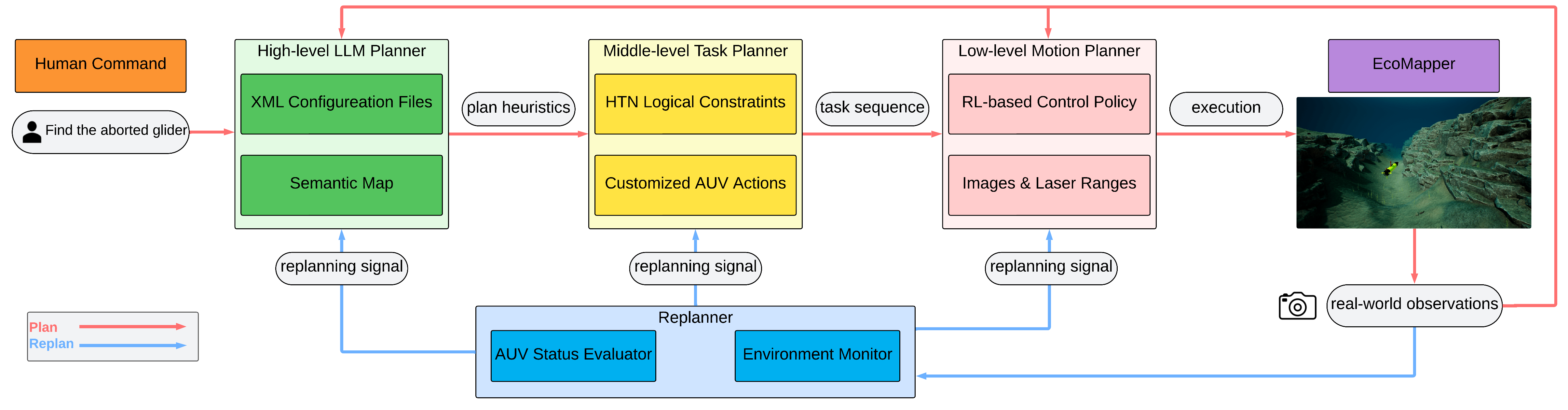}}
        \caption{Hierarchical framework of LLM-task-motion planning and replanning.}
        \label{framework}
\end{figure*}

The overall framework is supposed to bridge the gaps from an abstracted human command to physical AUV control while addressing specific challenges in the ocean. As shown in Figure~\ref{framework}, the framework is composed of four modules: i) A high-level LLM planner composes a plan heuristic of the human command to guide subsequent planning; ii) A middle-level task planner creates a feasible plan given the plan heuristic and predefined actions; iii) A low-level motion planner computes control inputs to execute the plan based on real-world observations; iv) A holistic replanner evaluates AUV status and reports unexpected situations to the corresponding planner for robust execution. This hierarchy decomposes planning complexity of the overall problem as high-level planners simplify planning for their low-level partners.

\subsection{Generalized LLM Planner}

The LLM planner retains the basic functionality of interpreting the command, but emphasizes more on its generalized knowledge to guide the task planner as a human-like brain, which offers two advantages: i) It can adapt into open ocean worlds based on its generalized inference rather than manually designed scene-specific prompts; ii) A reasonable plan heuristic efficiently biases search directions of achieving the human command in the large-scale ocean. Considering AUV challenges, we employ the following specific strategies.

\textbf{Semantic Map:} Identifiable objects can be sparse in the large-scale ocean, so the LLM planner must enrich its knowledge of the world given new observations. The LLM planner maintains an internal semantic map $\mathcal{M}_t$ to memorize the explored environment so far. We utilize Vision-Language Models (VLMs) to convert image observations into texts, so that the LLM planner will be progressively grounded in the mission and reinforce the future plan heuristic to avoid repeated exploration. For example when searching the warship, the AUV detects a glider next to it. In the future given a new command "Survey the glider", the LLM planner will prioritize the warship area.

\textbf{Augmenting XML File:} Unlike indoor robotic scenes, ocean environments may not possess sufficient semantic information for the LLM commonsense reasoning. We use an XML file $f_c$ to augment marine knowledge of the LLM planner. For instance, it can provide hints like "Coral reefs usually grow in areas with ample sunlight". The plan heuristic could be "The bright plain on the left is more likely to grow the coral reef than the dark hill on the right". Note that we still don't hands-on teach the LLM planner how to achieve AUV missions but just provide related marine knowledge.

We instantiate the current observation $z_t$ as agent-view RGB images of the surrounding environment. Then we use a VLM to associate each image with a text descriptor like "a rocky hill on the left". We score these text descriptors with the probability of how likely they will complete the command given the XML file and the current semantic map. The one with the highest probability is selected as the plan heuristics $h$. By providing query-response pairs with the LLM planner, the plan heuristic is eventually structured into a formal task consisting of symbolic initial state $s_{init}$ and goal state $s_{goal}$ for the task planner.

\subsection{HTN Task Planner}
\label{htn task planner}

We choose HTN planner \cite{nau1999shop} as the symbolic task planner and center the actions $\mathcal{A}$ on marine autonomy. For example, given the task "Survey the warship", the task planner should generate an action sequence $\Pi$ to guide the AUV to the warship without collision in the complex ocean. As such, we define the following Boolean-valued predicates:
\begin{itemize}
    \item \texttt{navigated ?auv - AUV}: if the AUV is navigated towards a certain area.
    \item \texttt{env\_sensed}: if a certain area is sensed.
    \item \texttt{detected ?target - object}: if the target object is detected.
    \item \texttt{captured ?target - object}: if the target object is taken photo of.
    \item \texttt{approached ?target - object}: if the target object is approached at a close distance.
    \item \texttt{reported ?target - object}: if photos of the target object is sent to the pilot.
    \item \texttt{replanned}: if the replanning signal is sent to the task planner.
\end{itemize}
The AUV actions are defined as follows:
\begin{itemize}
    \item \texttt{navigate(AUV)}: navigate the AUV towards a certain area.
    \item \texttt{sense}: take images of a certain area.
    \item \texttt{approach(target)}: move close to the target object.
    \item \texttt{capture(target)}: take photos of a target object.
    \item \texttt{report(target)}: surface to transmit photos and the GPS position of the target object.
    \item \texttt{rescue}: surface and send a rescue signal.
\end{itemize}
Detailed preconditions and effects of each action are presented in Appendix. We would like to delve deeper into the abstraction nature of our defined actions. There exist large-scale and unstructured landscapes like canyons and hills in the ocean, all of which pose collision risks to AUVs. However, these landscapes will not differ much across different ocean regions and may appear quite similar in high-dimensional representations like images. Furthermore, there are much less semantic objects in the ocean compared to indoor environments. In this sense, the unstructured ocean topography with sparse semantic information constitutes a challenging factor of controlling AUVs. By abstracting motions like "swim through a canyon" or "bypass a hill" into a unified action \texttt{navigate(AUV)}, we only need to capture the overall landscape around the AUV through visual observations. In this way, we sidestep impractical ocean map modeling or precise object detection along the AUV trajectory.

\subsection{DQN Motion Planner}

The motion planner determines control inputs for the AUV to execute in the physical world. As presented in Section \ref{problem motion planning}, we are motivated to leverage DRL methods to guide AUV motions based on these visual inputs. We select Deep Q Network (DQN) \cite{mnih2015human} as our solution, which aims at ﬁnding a direct mapping (represented by a DNN $\theta$) from the current observation $z_t$ to the Q-function $Q_\pi(z_t, u_t | \theta)$ and then uses greedy algorithm to generate a control policy
\begin{equation}
  \pi_{a_i}(u_t|z_t) \Leftarrow \underset{u_t}{\arg \max} Q_\pi(z_t, u_t | \theta)
\end{equation}
associated with an action $a_i \in \mathcal{A}$. The control input $u_t$ is drawn from the control policy $u_t \sim \pi_{a_i}(\cdot | z_t)$ given the current observation $z_t$. We instantiate $z_t$ as the agent-view RGB image and focus on learning the control policy associated with the \texttt{navigate(AUV)} action. The control policy plans a primitive motion like "move forward" or "turn left" given the current image, enabling the AUV to safely explore the specified region.

As mentioned in Section \ref{htn task planner}, we generalize the control policy of the \texttt{navigate(AUV)} action across diverse landscapes. By leveraging spatial representations embedded in images, the same control policy can be trained without differentiating between canyons or hills. DRL approaches typically take millions of steps to learn composite tasks. Our proposed planning hierarchy tremendously simplifies the motion planner into short-range movements. Moreover, our simulator HoloEco enables safe interaction with the environment so we can freely collect training datasets.

\subsection{Holistic Replanner}

Ocean environments are fairly uncertain and dynamic, so it is essential to adjust AUV behaviour through replanning. We design a holistic replanner to trigger replanning at respective planners by simultaneously considering the plan heuristic, action effects, and AUV states. This hierarchical replanning at all planners is an intuitive yet effective approach. Low-level issues like motion drift can be directly addressed by the motion planner without altering the entire plan. High-level problems should be addressed by the task planner injecting a corrective action or by the LLM planner re-assessing the unfinished mission. Given real-world feedback, the symbolic replanning flag is designed as follows:
\begin{equation}
\label{equation replanning}
    f = \left\{ 
    \begin{array}{lcl}
        \varnothing, &  &   \text{normal but mission not done} \\
        0,           &  &   \text{mission done} \\
        1,           &  &   \text{LLM replanning required} \\
        2,           &  &   \text{task replanning required} \\
        3,           &  &   \text{motion replanning required}
    \end{array}
    \right.
\end{equation}
We instantiate feedback to the replanner as three sensors: an IMU sensor, a forward laser sensor, and a velocity sensor. Every time the AUV executes the current control input $u_t$ in the real world, the updated observation $z'$ is obtained to analyze the numerical AUV state. Specifically, the replanner comprises two components: an AUV status evaluator and an environment monitor.

\textbf{AUV Status Evaluator:} 
The AUV status evaluator tracks abnormal AUV behavior. A significant reduction in AUV velocity indicates degeneration of the AUV mobility. With the replanning flag marked as $f = 2$, the task planner immediately terminates the current plan and implements the \texttt{rescue} action for assistance. If the IMU sensor detects radical AUV accelerations, the replanning flag will be triggered as $f = 3$, and the motion planner generates a corrective control input to restore the original moving direction.

\textbf{Environment Monitor:} 
The environment monitor keeps assessing the surroundings. If the AUV hasn't achieved the mission after the current plan heuristic, the replanning flag is set as $f = 1$. The LLM planner re-assesses the environment, updates the semantic map, and generates a new plan heuristic. Since the \texttt{navigate(AUV)} action is trained towards unstructured landscapes without accounting for objects, we employ a forward laser sensor in case of any collision in front of the AUV. With the replanning flag as $f = 3$, the motion planner controls the AUV away from the collision direction.

\section{Experiments}
\label{experiments}

We evaluate in HoloEco simulator that given one single human command, if the AUV can efficiently and safely navigate towards the target in  the large-scale unexplored ocean environment.


\subsection{Experimental Setup}
\label{experimental setup}

To evaluate our system, we build a marine simulator HoloEco upon HoloOcean \cite{Potokar22icra}, which provides high scalability and fidelity for AUV activities in a 3D ocean environment. It also ensures no damage to AUVs during the DQN training process. We instantiate three objects "coral reef, glider, warship" and three unstructured landscapes "canyon, hill, plain". We provide the VLM with reference images of objects and landscapes so that it can accurately interpret images for the LLM planner. We instantiate the AUV model as a precise prototype EcoMapper \cite{wang2015dynamic}. This detailed dynamics of the scene and the AUV presents how real missions can be similarly performed using our method.

\subsection{Mission 1 - Search Aborted Warship}

    \begin{figure*}
        \centerline{\includegraphics[width=\textwidth]{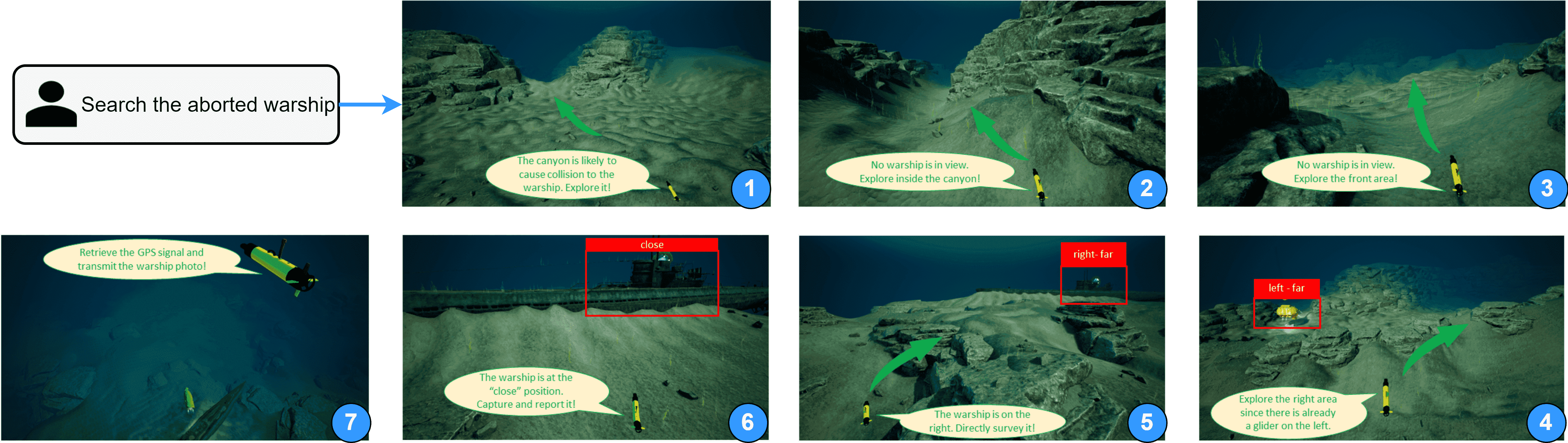}}
        \caption{The entire process of EcoMapper searching the aborted warship given an abstracted human command. Each numbered picture corresponds to a specific phase of the process.}
        \label{search aborted warship}
    \end{figure*}

After we issue an abstracted command "Search the aborted warship", OceanPlan directs EcoMapper to accomplish this comprehensive mission across the wide unknown ocean. The entire process is shown in Figure~\ref{search aborted warship}. Through seven phases of planning and replanning, EcoMapper successfully locates and reports the warship after exploring a wide range of the ocean. For example in phase 5, considering the command and the detected warship image, the LLM planner generates a plan heuristic "Directly survey the warship" and structures it into a formal task as 
\begin{itemize}
     \item  \texttt{s\_{init} = (not (approached warship)) (detected warship) (not (reported warship)) }

     \item  \texttt{s\_{goal} = (reported warship)} 
 \end{itemize}
To achieve the task, the HTN task planner formulates a plan \texttt{[approach(warship), capture(warship), report(warship)]} and the motion planner sequentially executes the actions in the plan. Once the current plan is executed, the replanner evaluates both mission progress and the AUV status. A full video is available at the project website \url{https://sites.google.com/view/oceanplan}.

\subsection{Mission 2 - Search Glider near Coral Reef}

In this mission, there is a glider working around the coral reef. The AUV pilot would like to check the glider by issuing an abstracted command "Search the glider near coral reef". In phase 1, the VLM detects a warship and a glider. The LLM planner updates the semantic map with "A glider is near the warship on the right". The plan heuristic is "Search the left area for the other glider near the coral reef". In phase 2 with few objects, environment semantic information is extremely sparse. The LLM planner turns to the XML file in Section \ref{experimental setup} for more hints and generates a plan heuristic "Based on the coral reef attribute in the XML file, it is likely to grow in the front plain with ample sunlight". A full video is available at the project website \url{https://sites.google.com/view/oceanplan}.


\subsection{Mission 3 - Replan}

Through three unexpected situations where replanning is initiated at the corresponding planner, we present robust AUV operation in the unpredictable ocean. In Mission 1, replanning at the LLM planner has been extensively presented, so we focus on presenting replanning at task planner and motion planner. A full video is available at the project website \url{https://sites.google.com/view/oceanplan}.    

\textbf{Situation 1:} The replanner detects a significant reduction in AUV speed and sends a replanning signal to the task planner. The task planner immediately replaces the current plan with the \texttt{rescue} action so that EcoMapper surfaces and transmits the rescue signal in time.

\textbf{Situation 2:} The replanner identifies a radical leftward acceleration of AUV, triggering the replanning flag of the motion planner. The motion planner responds with a corrective control input "turn right" to offset the unexpected leftward acceleration.

\textbf{Situation 3:} The replanner detects that EcoMapper is at a high risk of colliding with a glider in front of it. The motion planner controls EcoMapper to move away from the glider.

\subsection{Ablation Studies}

We perform two ablation studies of Mission 1 and Mission 2 to evaluate importance of LLM planner and task planner during long missions in the unexplored ocean. We carry out 10 simulation runs and average both the completion time and the success rate. We exclude response time of LLM and VLM. We aim to claim through quantitative comparison that lack of any component in our proposed framework will largely depreciate the AUV performance. 

\begin{figure}
        \centerline{\includegraphics[width=0.48\textwidth, height=5cm]{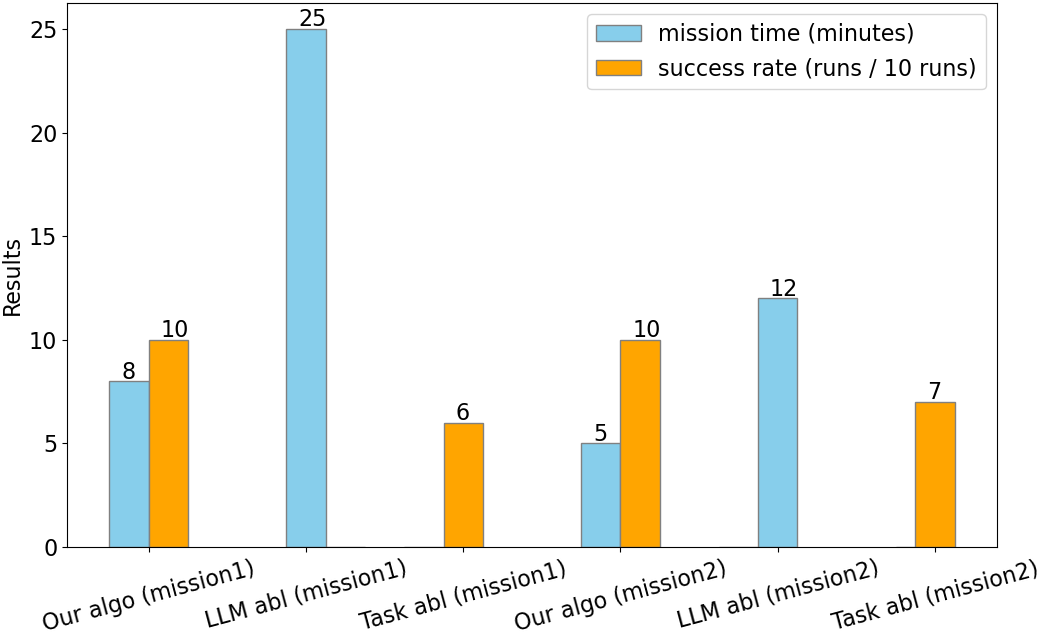}}
        \caption{Quantitative results of ablation studies demonstrate that our method achieves a good balance between efficiency and validity of planning a long-horizon mission given an abstracted command.}
        \label{ablation}
\end{figure}

\textbf{Ablation of LLM Planner:}
In this study, we only use task planner and motion planner to achieve the same command with the same actions. In each phase, we manually evaluate the images and form a task for the task planner. We only evaluate if the target object is in view and don't consider semantic information about other objects. If not in view, we randomly choose an exploring direction. As shown in Figure~\ref{ablation}, it takes around three times longer than the proposed method to achieve Mission 1 and around twice to achieve Mission 2. We can conclude that it is inefficient to rely solely on the task planner and motion planner to perform long-horizon planning given an abstracted command. In absence of heuristics from the LLM planner, the task planner will take much more time to randomly explore the ocean space in a brute-force pattern.

\textbf{Ablation of Task Planner:}
In this study, we only use LLM planner and motion planner. We provide the LLM planner with the same actions, but don't provide their preconditions and effects or illustrative prompts. The LLM planner relies on itself to organize the actions to achieve the heuristic. As shown in Figure~\ref{ablation}, the LLM planner generates invalid plans 6 out of 10 runs in Mission 1 and 7 out of 10 runs in Mission 2. For example in Mission 1, for the heuristic "Directly survey the warship", a wrong plan \texttt{[navigate(ecomapper), sense, approach(warship), navigate(ecomapper), sense, capture(warship), navigate(ecomapper), sense, report(warship)]} is generated, where the second \texttt{navigate(ecomapper)} action causes EcoMapper to collide with the warship. We can conclude that without logical connections introduced by the task planner, the LLM planner cannot guarantee the plan quality.

\section{Conclusion}
\label{conclusion}

We propose a hierarchical planning and replanning framework to pilot AUVs through natural language in the large-scale unknown ocean given major marine robotics challenges. Given a human command, an LLM planner generates a plan heuristic, a task planner guarantees a valid plan, a motion planner executes the plan in the real world, and a holistic replanner ensures robust AUV operation. In a marine simulator HoloEco, OceanPlan is validated to ground human commands for effective and safe AUV missions.



\newpage

\bibliography{IEEEabrv, reference.bib}
\bibliographystyle{IEEEtran}

\appendix

\section*{Action Preconditions and Effects}
\begin{figure}
        \centerline{\includegraphics[width=0.48\textwidth]{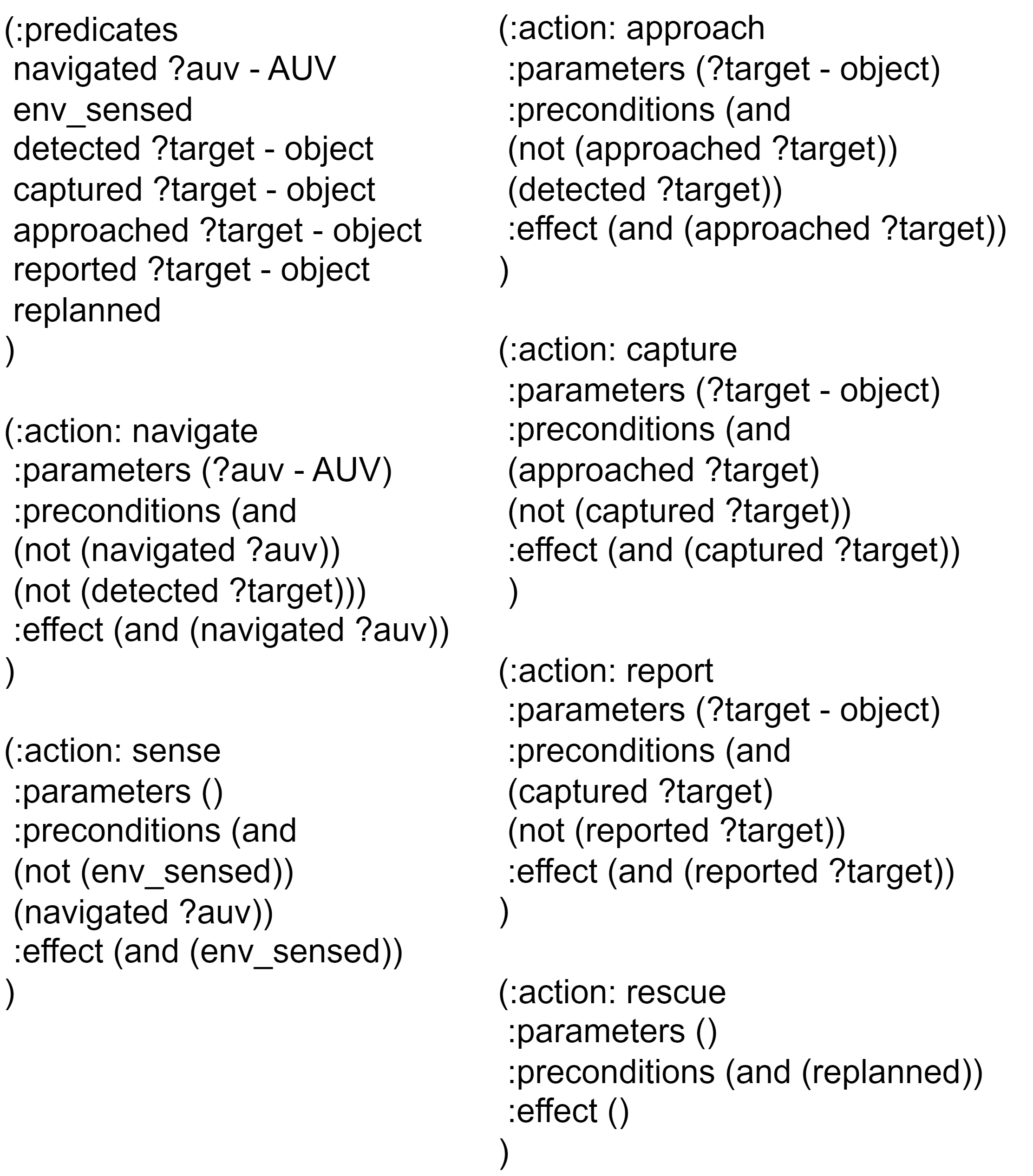}}
        \caption{Preconditions and effects of predefined AUV actions.}
        \label{predefined actions}
\end{figure}

Detailed preconditions and effects of predefined actions are presented in Figure~\ref{predefined actions}.

\section*{Action Implementation}

We illustrate detailed implementation of the control policies associated with the \texttt{navigate(AUV)} and \texttt{approach(target)} actions. 

\subsection{\texttt{navigate(AUV)} action}

\begin{figure}
     \centering
     \hfill
     \begin{subfigure}{0.49\columnwidth}
         \centerline{\includegraphics[width=\linewidth]{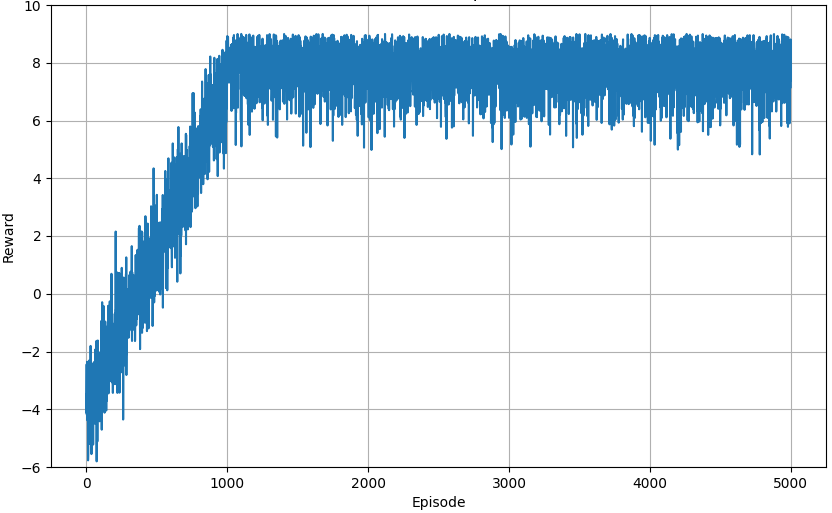}}
          \caption{Reward over episode.}
         \label{reward}
     \end{subfigure}   
     \hfill
     \begin{subfigure}{0.49\columnwidth}
         \centerline{\includegraphics[width=\linewidth]{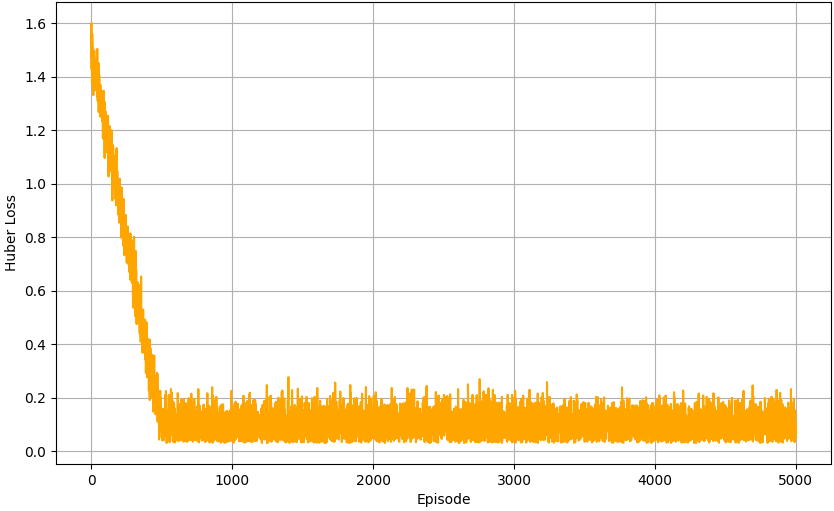}}
         \caption{Huber loss over episode.}
         \label{loss}
     \end{subfigure} 
    \caption{DQN training results.}
    \label{DQN training result}
\end{figure}

The control policy of the \texttt{navigate(AUV)} action is a learned stochastic policy function which takes the current image as input and outputs a primitive action. We utilize DQN to train the control policy, which requires training two policies: the behaviour policy and the target policy with the same network parameters. The behaviour policy collects experiences through interaction with the simulated environment, while the target policy learns from them to update its own network. We follow the training pipeline of \cite{devo2020towards} and instantiate the training scenario as EcoMapper navigating through a canyon, hill, and plain safely and quickly. 
\begin{itemize}

    \item Control inputs: turn left, turn right, move forward, move up.

    \item Observation: the current agent-view RGB image.

    \item Immediate reward: time penalty -0.1 to encourage shorter trajectories; collision penalty -0.1 to discourage collision. Goal-reaching reward +10 upon action completion.

    \item Terminated condition: The AUV safely navigates through the area. Truncated condition: Control inputs exceed 30. 

     \item  Network architecture: We use Convolutional Neural Network to extract features of the current image. Next we flatten and feed the features to a Feedforward Network, which returns the Q value of all control inputs.
     
\end{itemize}
The training results in Figure~\ref{DQN training result} present convergence of both the loss and the reward. The training hyperparameters are shown in Table~\ref{hyperparameters}.

\begin{table}  
\caption{DQN training hyperparameters.} 
\begin{center}
\label{hyperparameters}
\begin{tabular}{| c | c |}
\hline
Hyperparameters  &  Value  \\
\hline
discount factor   & 0.95    \\
replay memory total size   & 60000  \\
relay memory batch size    & 64    \\
learning rate    & 0.005    \\
initial training samples  & 2000    \\
target policy updating frequency     & 500   \\
training episodes    & 5000   \\
epsilon limit of behavior policy & 0.05 \\
\hline
\end{tabular}
\end{center}
\end{table}


\subsection{\texttt{approach(target)} action} 

\begin{figure}
        \centerline{\includegraphics[width=0.48\textwidth]{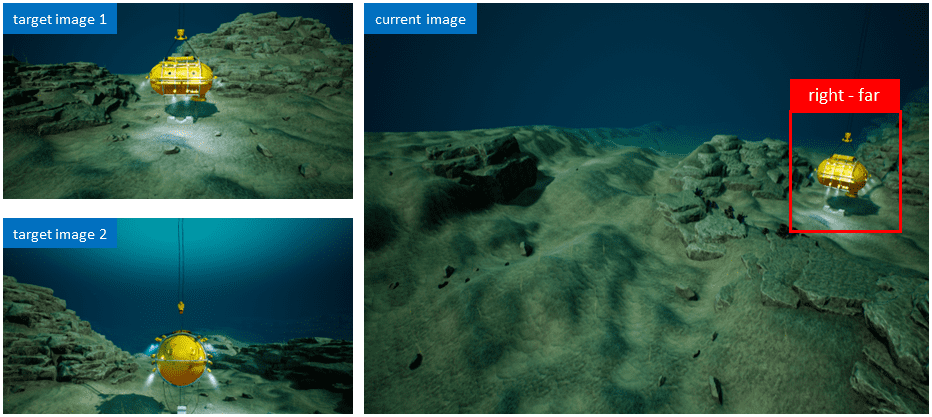}}
        \caption{On the left side, two target images show the target object 'glider' at the 'center-close' position. On the right side, the current image is identified with the 'glider' at the 'right-far' position.}
        \label{approach action}
    \end{figure}

The control policy of the \texttt{approach(target)} action is to move the AUV close to the target object by tracking the target in the image center. The VLM outputs the relative position of the target with respect to the AUV by comparing the current image and the target images. The VLM first identifies if the target object is 'close' or 'far' to the AUV. If deemed 'far', the position is discretized into five categories: 'left-far', 'right-far', 'center-far', 'top-far', and 'bottom-far'. We take into account three classes of objects 'glider', 'warship', and 'coral reef'. Following question-answer samples, the VLM can identify the object's relative position in the image. The \texttt{approach(target)} action will not end until the target is detected as 'close'. Since there is no training for this action, we use a forward laser sensor to avoid collision with the target object. An illustrative detection result is shown in Figure~\ref{approach action}.

\end{document}